# An attempt to generate new bridge types from latent space of energy-based model

Hongjun Zhang

Wanshi Antecedence Digital Intelligence Traffic Technology Co., Ltd, Nanjing, 210016, China

583304953@QQ.com

**Abstract:** Use energy-based model for bridge-type innovation. The loss function is explained by the game theory, the logic is clear and the formula is simple and clear. Thus avoid the use of maximum likelihood estimation to explain the loss function and eliminate the need for Monte Carlo methods to solve the normalized denominator. Assuming that the bridge-type population follows a Boltzmann distribution, a neural network is constructed to represent the energy function. Use Langevin dynamics technology to generate a new sample with low energy value, thus a generative model of bridge-type based on energy is established. Train energy function on symmetric structured image dataset of three span beam bridge, arch bridge, cable-stayed bridge, and suspension bridge to accurately calculate the energy values of real and fake samples. Sampling from latent space, using gradient descent algorithm, the energy function transforms the sampling points into low energy score samples, thereby generating new bridge types different from the dataset. Due to unstable and slow training in this attempt, the possibility of generating new bridge types is rare and the image definition of generated images is low.

## 0  Introduction

Common innovative methods include combination, analogy, and enumeration, all of which have driven technological progress in bridge engineering. For example: ①Tied-arch bridge is a combination of beam bridge and arch bridge; ②French horticulturist Joseph Monier observed that plant roots can improve the crack resistance of soil and invented reinforced concrete through analogical thinking; ③Bridges are constructed using stronger concrete and steel instead of the original wood and stone materials, which can be classified as the attribute enumeration method.

Only using bridge image datasets to train a neural network model can only achieve geometric combination innovation of bridge types, and cannot achieve innovative methods such as analogy and enumeration. Using image datasets with multiple structures and objects, or their pre trained models, or even multimodal models, can bring a new perspective and enable innovation in analogy, enumeration, and other methods. However, this requires higher requirements for algorithms, data, hardware, and more.

The Energy Based Model (EBM) [1-4] used here, similar to the Variational Autoencoder (VAE), Generative Adversarial Network (GAN), Pixel Convolutional Neural Network (PixelCNN), and Normalizing Flow (NF) used by the author in previous papers [5-8], only use bridge image dataset, train neural network model to achieve geometric combination innovation of bridge types.

The energy-based generative model also maintains the number of dimensions in the sample space unchanged (similar to pixel convolutional neural network and normalizing flow). It constructs a neural network to represent the energy function and trains it to accurately calculate the energy values of real and fake samples (real samples have low energy values, while fake samples have high energy values). Then, samples are taken from the latent space and gradient descent algorithm is used to transform the sampling points into low energy value samples using the energy function, thereby obtaining realistic generated images.

This article establishes an energy-based model, using the same bridge image dataset as before, and further attempts geometric combination innovation of bridge types (open source address of this article's dataset and source code: https://github.com/QQ583304953/Bridge-EBM).

# 1 Introduction to energy-based model
## 1.1 Overview

The core idea of the energy-based models draw inspiration from the probability model of particle distribution in the potential field of statistical physics (Boltzmann distribution law). When there is a conservative external force (such as a gravitational field), the ideal gas particles are no longer uniformly distributed, and the particle density varies at different positions in space. The particle distribution law is described using the Boltzmann distribution law.

The earliest application of energy-based models in machine learning was the Boltzmann Machine in 1985, whose main idea was to use energy to learn unknown probability distribution. For a given dataset, it is very difficult to learn how to directly write the likelihood function without knowing its distribution form. And various probability distributions can be transformed into each other (the distribution of random variable function). By appropriate transformation, the unknown probability distribution of the dataset can be transformed into a Boltzmann distribution. So an energy-based model is a method for learning the unknown probability distribution of a dataset.

## 1.2 Boltzmann distribution

Taking ideal gases in a gravitational field as an example. Gas particles are subjected to two effects: their thermal motion causes them to tend towards a uniform distribution in space, while gravity forces them to tend towards falling towards the ground. When the two effects reach equilibrium, the particle density and pressure of the gas will decrease with increasing height. In reality, the air density on the ground is much higher than that at high altitude.

1. The pressure of the gas is:

$$p = nkT \tag{1}$$

In the formula: p is the pressure of the gas, n is the particle number density, k is the physical constant, and T is the temperature (the temperature of the gas in equilibrium is the same everywhere).

2. Cut a vertical cylinder in the gas, and according to fluid statics, the difference between the pressure p at height h and the pressure p' at height h+$dh$ in the stationary gas is:

$$dp = p' - p = -\rho g dh = -nmg dh \tag{2}$$

In the formula: $\rho$ is the mass density of the gas, g is the gravitational acceleration, and m is the mass of a single particle.

3. Due to the same temperature throughout the gas, this pressure difference can only be caused by differences in particle number density n, i.e.:

$$dp = kT dn \tag{3}$$

From equations (1) and (2), it can be concluded that:

$$-nmg dh = kT dn \tag{4}$$

$$\frac{1}{n} dn = -\frac{mg}{kT} dh \tag{5}$$

$$\int \frac{1}{n} dn = -\frac{mg}{kT} \int dh \tag{6}$$

$$\ln|n| = -\frac{mg}{kT} h + C \tag{7}$$

$$n = e^C e^{-\frac{mgh}{kT}} \tag{8}$$

$$= n_0 e^{-\frac{\varepsilon}{kT}} \tag{9}$$

In the formula: $n_0$ is the particle number density when h=0, $\varepsilon$ The gravitational potential energy of the particle mgh.

4. Formulas (8) and (9) indicate that particles in the potential field always preferentially occupy positions with lower potential energy.

In space, there are areas with high energy and areas with low energy. The lower the energy, the greater the probability of particles appearing.

The distribution pattern of particles in space is described by the Boltzmann distribution law:

$$p(x) = \frac{e^{-E(x)}}{Z} \quad (10)$$

In the formula: x is the spatial position, E(x) is the energy function (energy is determined by spatial position), Z is the normalized denominator $\int e^{-E(x)} dx$ (the integration interval is the entire space).

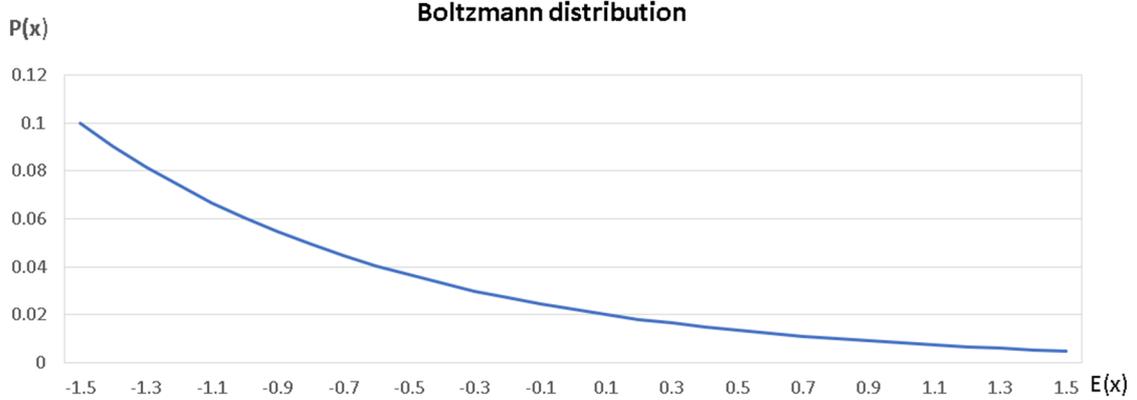

Fig.1 Boltzmann distribution

When the probability of an event is expressed as a Boltzmann distribution, x in equation (10) represents the event.

The samples and sample features in the energy-based model correspond to the particles and their spatial positions in equation (10).

### 1.3 Sample space and latent space

The sample space of images is a high-dimensional pixel space, with complex correlations between dimensions. The latent space is artificially designed, which is the hypothesis space of the sample space transformed by a neural network. There is a mapping relationship between the latent space and the sample space.

The latent space dimensions of energy-based models are independent of each other, with the same number of dimensions as the sample space, and the dimensions correspond one-to-one. Latent space coordinates are the initial seeds for generating samples. The latent space coordinate values can be taken as floating-point values from 0 to 1, while the sample space coordinate values are generally integer values from 0 to 255.

### 1.4 Loss function

1. Reference [9] uses the maximum likelihood estimation method to establish the loss function, and the Monte Carlo method approximates the normalized denominator $\int e^{-E(x)} dx$. Details can be found in A Appendix.

The logic of this method is not clear enough, and the formula form is not intuitive enough.

2. This article adopts the concept of game theory to explain the loss function, with clear logic and concise formulas. Specifically, as follows:

(1) Most of the sample points in the sample space have no practical significance, or in other words, the probability tends to zero. The dataset samples are all from practice and have practical significance, and their corresponding sample points have a clear probability of occurrence. Assuming the population follows a Boltzmann distribution, the dataset must be located in the region with low energy.

(2) So when inputting the dataset samples (real samples) into the energy function, low energy values should be output. Random images and images with completely different features from the dataset (fake samples) should output high energy values. Therefore, the loss function is constructed as follows:

$$\text{Loss}(\theta) = \mathbb{E}[E_\theta(x)] - \mathbb{E}[E_\theta(y)] \quad (11)$$

In the formula: $\theta$ The weight parameters of the energy function(neural network), $\mathbb{E}[E_\theta(x)]$ is the energy mean of the dataset samples, $\mathbb{E}[E_\theta(y)]$ The energy mean of fake samples.

(3) If fake samples are directly taken from images with completely different features in the dataset, as the training progresses, the loss values will tend to negative infinity. The neural network only needs to learn to classify based on individual feature in the dataset, which is obviously meaningless.

Therefore, fake images are generated by energy function, which will contain the dataset features learned by the previous epoch of neural network. This forces the neural network to learn more dataset features in the next epoch of training. Repeat this process until the neural network has mastered all the features of the dataset and the loss value approaches 0. At this point, the fake image generated by the energy function will be very realistic.

### 1.5 Generating samples with energy function

1. Given an image input, the energy function outputs an energy value scalar. How can we use this function to generate a new sample with low energy value? The method is to use a technique called Langevin dynamics[3]. Firstly, starting from random points in the latent space, calculate the gradient of the energy function output relative to the image input, then update the image input slightly in the negative gradient direction, and then input it to the energy function. Through such multiple iterations, random points will gradually transform into images with dataset sample features. The entire process maintains the weight parameters of the energy function unchanged. To avoid getting stuck in local minima, add a small amount of noise to the image input before inputting it to the energy function[10].

$$x_{new} = x_{old} - \alpha * \nabla_x E(x) \qquad (12)$$

In the formula: $x_{new}$ is the updated image, $x_{old}$ is the image before the update, $\alpha$ is the step size (learning rate), $E(x)$ is the energy function (energy is determined by image features).

2. The energy diagram of the samples can be simulated and described using a terrain with valleys and ridges intersecting.

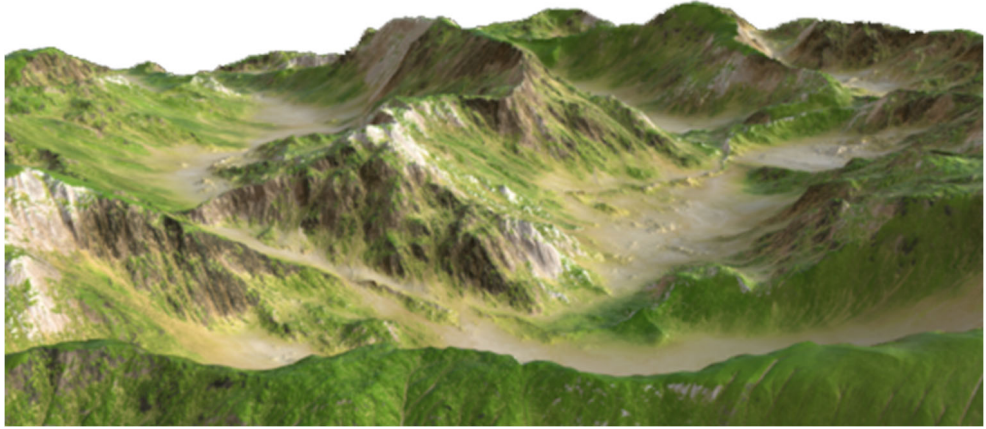

Fig.2 Energy diagram of the samples

The XY horizontal plane is assumed to represent all dimensions of the samples, and the Z-axis is assumed to represent energy values. Samples with similar features will gather together. The dataset samples are distributed at the bottom of the valleys (areas with low energy), while at the ridges (areas with high energy) there are random noises or irrelevant images. The samples generated by latent space sampling are closer to valleys, lower in energy, and more similar to dataset samples, with higher clarity; Samples located close to the halfway up the mountain often have the characteristics of two types of samples, which may achieve geometric combination, but clarity and practical significance are difficult to guarantee, and there is a great deal of randomness.

## 2 An attempt to generate new bridge types from latent space of energy-based model
### 2.1 Dataset

Using the dataset from the author's previous paper [5-8], which includes two subcategories for each type of bridge (namely equal cross-section beam bridge, V-shaped pier rigid frame beam bridge, top-bearing arch bridge, bottom-bearing arch bridge, harp cable-stayed bridge, fan cable-stayed bridge, vertical_sling suspension bridge, and diagonal_sling suspension bridge), and all are three spans (beam bridge is 80+140+80m, while other bridge types are 67+166+67m), and are structurally symmetrical.

This model is difficult to train, so the image size is reduced from 512x128 to 192x48 pixels (the disadvantage is that the clarity is much reduced).

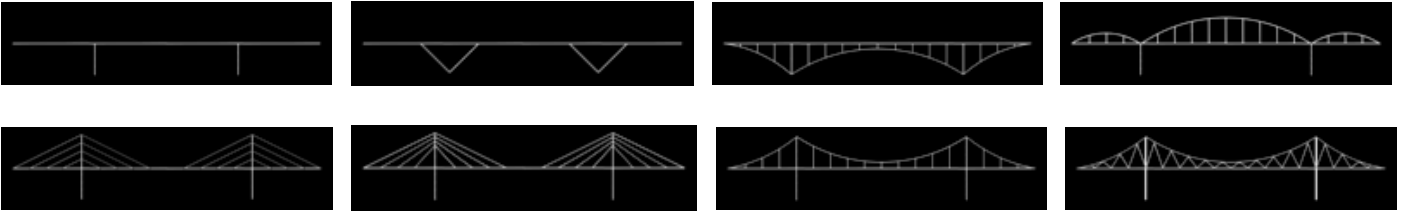

Fig.3 Grayscale image of each bridge facade

Each sub bridge type obtained 1200 different images, resulting in a total of 9600 images in the entire dataset.

## 2.2 Construction and training of energy-based model

Based on the Python3.10 programming language, TensorFlow2.10, and Keras2.10 deep learning platform framework, construct and train EBM[10].

1. Architecture of energy function

The job of energy function is to calculate the energy values of the input images.

First, create an input layer for the images, then passed to four Conv2D layers in sequence (activation function swish), and finally flatten and connect to two Dense layers in sequence (activation function swish and linear transformation). The output result of every image is scalar. The model summary is shown in the table below:

Tab.1 model summary of energy function

| Layer (type) | Output Shape | Param # |
| --- | --- | --- |
| input_1(InputLayer) | [(None,48,192,1)] | 0 |
| conv2d(Conv2D) | (None,24,96,32) | 832 |
| batch_normalization(BatchNormalization) | (None,24,96,32) | 128 |
| activation(Activation) | (None,24,96,32) | 0 |
| conv2d_1(Conv2D) | (None,12,48,64) | 18496 |
| batch_normalization_1(BatchNormalization) | (None,12,48,64) | 256 |
| activation_1(Activation) | (None,12,48,64) | 0 |
| conv2d_2(Conv2D) | (None,6,24,128) | 73856 |
| batch_normalization_2(BatchNormalization) | (None,6,24,128) | 512 |
| activation_2(Activation) | (None,6,24,128) | 0 |
| conv2d_3(Conv2D) | (None,3,12,128) | 147584 |
| batch_normalization_3(BatchNormalization) | (None,3,12,128) | 512 |
| activation_3(Activation) | (None,3,12,128) | 0 |
| flatten(Flatten) | (None,4608) | 0 |
| dense(Dense) | (None,128) | 589952 |
| dense_1(Dense) | (None,1) | 129 |
| Total params: 832,257 | | |
| Trainable params: 831,553 | | |
| Non-trainable params: 704 | | |

2. Loss function

Energy models are not easy to train, and if hyperparameters are not well adjusted, they often diverge. Therefore, in addition to the difference in energy values between true and fake samples(Constractive Divergence,CD), regularization loss is also added to facilitate training.

Regularization loss is the mean of the sum of squared energy values of true and fake samples, multiplied by a weight factor (around 0.1). In this way, the energy values of true and fake samples tend to be negative and positive respectively, and the absolute values are basically the same.

3. Training

Generating fake samples from the model requires a significant amount of iterative computation. There is a training technique that can significantly reduce sampling costs: using sampling buffer. Store the fake samples from the previous batchs in the buffer and reuse them as the starting point for the next batch's sampling generation, so that fewer steps can result in higher quality samples. In order to not completely rely on previous samples, 5% of the samples in each batch are generated from scratch.

The loss variation curve during the training process is shown in the following figure (excluding the results of the first three rounds):

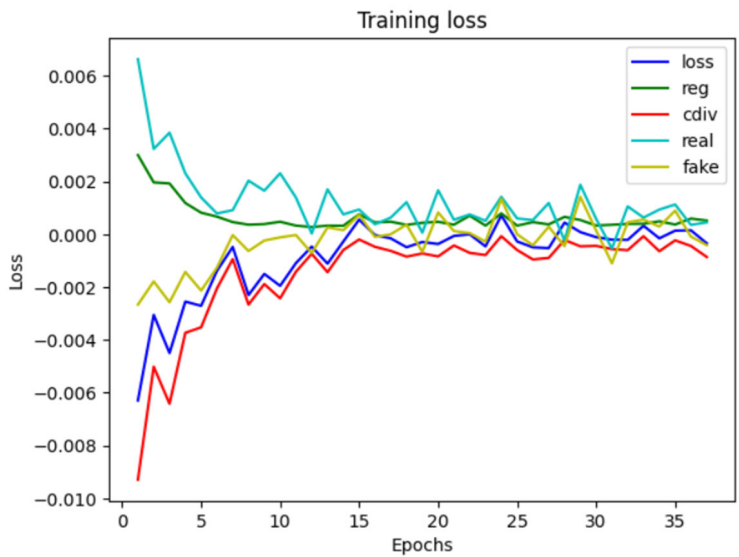

Fig.4 Training loss curve

Among them: loss is the total loss mean, Reg is the regularization loss mean, cdiv is the constractive divergence loss mean, real is the energy mean of the dataset samples, and fake is the energy mean of fake samples.

There are issues with unstable and slow training in this attempt, resulting in a rare possibility of generating new bridge types and low image clarity. Due to the constraints of hardware conditions, it is impossible to fully optimize. Thus, I can only test parameters while sampling.

4. The process of generating images from noise

To gain a deeper understanding of how images change during the iteration process, you can look at intermediate samples.

The following figure shows how a random point in latent space is gradually transformed into an image.

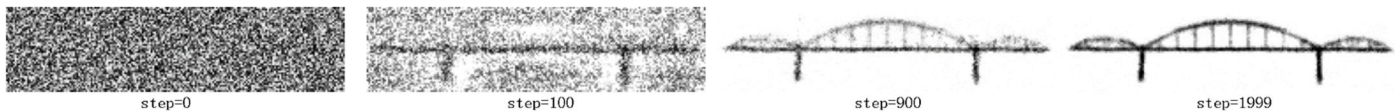

Fig.5 The process of generating images from noise

We can see that although the starting point starts with noise, the basic shape is obtained after 900 steps, and by step 1999, the shape becomes clear.

## 2.3 Exploring new bridge types through latent space sampling

Although some models were saved during training, the generated image quality and clarity are too low. During training, the image clarity in the sampling buffer is relatively high, so three technically feasible new bridge types are obtained through manual screening based on the thinking of engineering structure, which are completely different from the dataset:

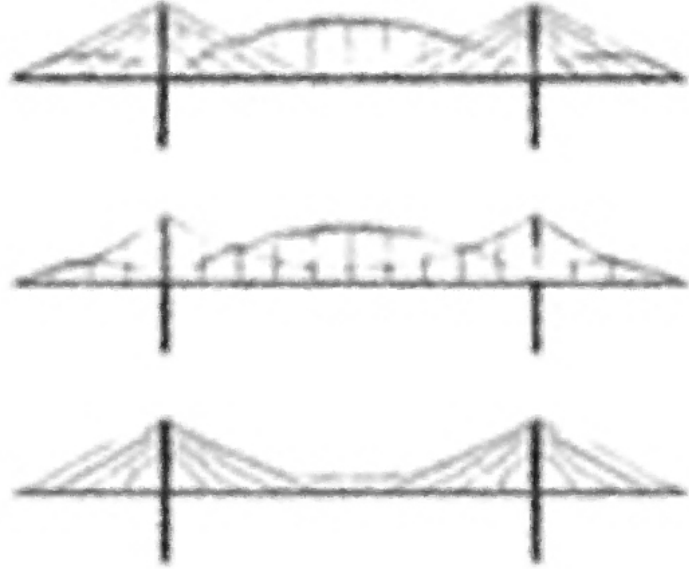

Fig.6 Three new bridge types with feasible technology

The new bridge type here refers to a type of bridge that has never appeared in the dataset, but is created by neural network based on algorithms, which represents the model's innovative ability.

## 3 Conclusion

Energy-based models can combine two types of bridge types based on human original bridge types, and create new bridge types by combining them.

Due to hardware constraints, the model parameters are not fully optimized, resulting in a rare possibility of generating new bridge types and low image clarity.

Using the game theory to explain the loss function, the logic is clear and the formula is concise, avoiding the use of maximum likelihood estimation to explain the loss function, and there is no need for Monte Carlo method to solve the normalized denominator, which greatly facilitates the mastery of the algorithm by ordinary people.

# A Appendix: Details of Section 1.4.1

## A.1 Loss function

The dataset is sampled from the population, so when training a neural network (energy function) with dataset samples $x_i$, the maximum likelihood estimation method can be used to establish a loss function [9]:

$$\text{Loss}(\theta) = -\frac{1}{n}\sum_{i=1}^{n} \ln p_\theta(x_i) \tag{13}$$

$$= -\mathbb{E}[\ln p_\theta(x_i)] \tag{14}$$

$$= \mathbb{E}[E_\theta(x_i)] + \ln z_\theta \tag{15}$$

Because it is a generative model, the goal is to give the energy function the ability to generate realistic images, so the normalized denominator $z_\theta$ is calculated using the energy value of the generated image, so as to achieve the goal.

To simplify the calculation, take the derivative of $\theta$ on both sides of the above equation:

$$\nabla_\theta \text{Loss}(\theta) = \mathbb{E}[\nabla E_\theta(x_i)] + \nabla \ln z_\theta \tag{16}$$

$$= \mathbb{E}[\nabla E_\theta(x_i)] + \frac{\nabla z_\theta}{z_\theta} \tag{17}$$

$$= \mathbb{E}[\nabla E_\theta(x_i)] - \frac{1}{z_\theta}\int e^{-E_\theta(x)} \nabla E_\theta(x)\, dx \tag{18}$$

$$= \mathbb{E}[\nabla E_\theta(x_i)] - \int \frac{e^{-E_\theta(x)}}{z_\theta} \nabla E_\theta(x)\, dx \tag{19}$$

$$= \mathbb{E}[\nabla E_\theta(x_i)] - \int p_\theta(x) \nabla E_\theta(x)\, dx \tag{20}$$

$$= \mathbb{E}[\nabla E_\theta(x_i)] - \mathbb{E}[\nabla E_\theta(x)] \tag{21}$$

$$= \nabla\{\mathbb{E}[E_\theta(x_i)] - \mathbb{E}[E_\theta(x)]\} \tag{22}$$

Note: For brevity, $\nabla_\theta$ is shortened to $\nabla$.

Integral $\theta$ on both sides of the above formula, ignoring the constant term:

$$\text{Loss}(\theta) = \mathbb{E}[E_\theta(x_i)] - \mathbb{E}[E_\theta(x)] \tag{23}$$

In the formula: $\mathbb{E}[E_\theta(x_i)]$ is the energy mean of the dataset sample $x_i$, $\mathbb{E}[E_\theta(x)]$ is the energy mean of the samples generated by model sampling. The Monte Carlo method can be used to approximate the solution of $\mathbb{E}[E_\theta(x)]$.

## A.2 Monte Carlo method

The Monte Carlo method is a statistical simulation method that approximates problems such as expected values, area, and integration through a large number of samples. For example, when the prior probability of a coin facing upwards is unknown, it can be approximately estimated by tossing it multiple times and then counting its frequency of facing upwards.

A batch of data points can be randomly sampled in the latent space and then transformed into samples $x_j$ that approximately follows the population distribution, so as to approximately solve $\mathbb{E}[E_\theta(x)]$.

$$\mathbb{E}[E_\theta(x)] \approx \frac{1}{J}\sum_{j=1}^{J} E_\theta(x_j) \tag{24}$$